\documentclass{article} % For LaTeX2e
\usepackage{iclr2023_conference_tinypaper,times}

\usepackage{amsmath,amsfonts,bm}

\def\eqref#1{equation~\ref{#1}}
\def\1{\bm{1}}

\DeclareMathAlphabet{\mathsfit}{\encodingdefault}{\sfdefault}{m}{sl}
\SetMathAlphabet{\mathsfit}{bold}{\encodingdefault}{\sfdefault}{bx}{n}

\usepackage{hyperref}
\usepackage{url}
\usepackage{graphicx}
\usepackage{tabularx}
\usepackage{longtable}
\usepackage{subcaption}

\title{lamper: language model and prompt engineering for zero-shot time series classification}

\author{Zhicheng Du$^1$, Zhaotian Xie$^1$, Yan Tong$^2$, Peiwu Qin$^{1,}\thanks{Corresponding author}$ \\
$^1$Tsinghua Shenzhen International Graduate School, Tsinghua University \\
$^2$School of Automation, Wuhan University of Technology \\
\texttt{\{duzc21, xzt21\}@mails.tsinghua.edu.cn, ty1909674582@outlook.com}\\ 
\texttt{pwqin@sz.tsinghua.edu.cn}\\ 
}

\iclrfinalcopy % Uncomment for camera-ready version, but NOT for submission.
\begin{document}

\maketitle
\begin{abstract}
This study constructs the LanguAge Model with Prompt EngineeRing (LAMPER) framework, designed to systematically evaluate the adaptability of pre-trained language models (PLMs) in accommodating diverse prompts and their integration in zero-shot time series (TS) classification. We deploy LAMPER in experimental assessments using 128 univariate TS datasets sourced from the UCR archive. Our findings indicate that the feature representation capacity of LAMPER is influenced by the maximum input token threshold imposed by PLMs. 
\end{abstract}

\section{Introduction}
\label{int}
The exploration of time series (TS)-based tasks constitutes a research-intensive domain with significant implications with wide-ranging implications in diverse professional fields, including healthcare, finance, and energy \citep{zhang2022optimising, ZHENG2023113046, santoro2023higher}. Within the realms of natural language processing (NLP), the dynamic landscape witnesses the rapid evolution of pre-trained language models (PLMs) and prompt engineering \citep{plmsurvey, wei2022finetuned}. These advancements underscore their commendable capacity to adeptly execute an extensive array of tasks, particularly under few-shot or even zero-shot conditions \citep{NEURIPS2020_1457c0d6, webson2022promptbased}. In stark contrast, various TS-based tasks necessitate substantial domain expertise and bespoke model design, thereby constraining both model performance and generalization capabilities. Recent efforts have seen the application of PLMs and prompt engineering to diverse TS tasks, encompassing forecasting, restoration, and anomaly detection \citep{tsbert, hu2023bertpin, jin2023timellm}. However, the effectiveness of amalgamating PLMs and prompt engineering for the realization of zero-shot learning in the realm of TS remains uncharted territory. In response, this study introduces LanguAge Models with Prompt EngineeRing (LAMPER) for exploring the latent potential of PLMs in feature representation for TS data, with a specific focus on achieving zero-shot TS classification through the strategic utilization of diverse prompts and their fusion.

\section{Method}
\label{met}
\begin{figure}[ht]
% \begin{center}
%\framebox[4.0in]{$\;$}
\includegraphics[width=1\linewidth]{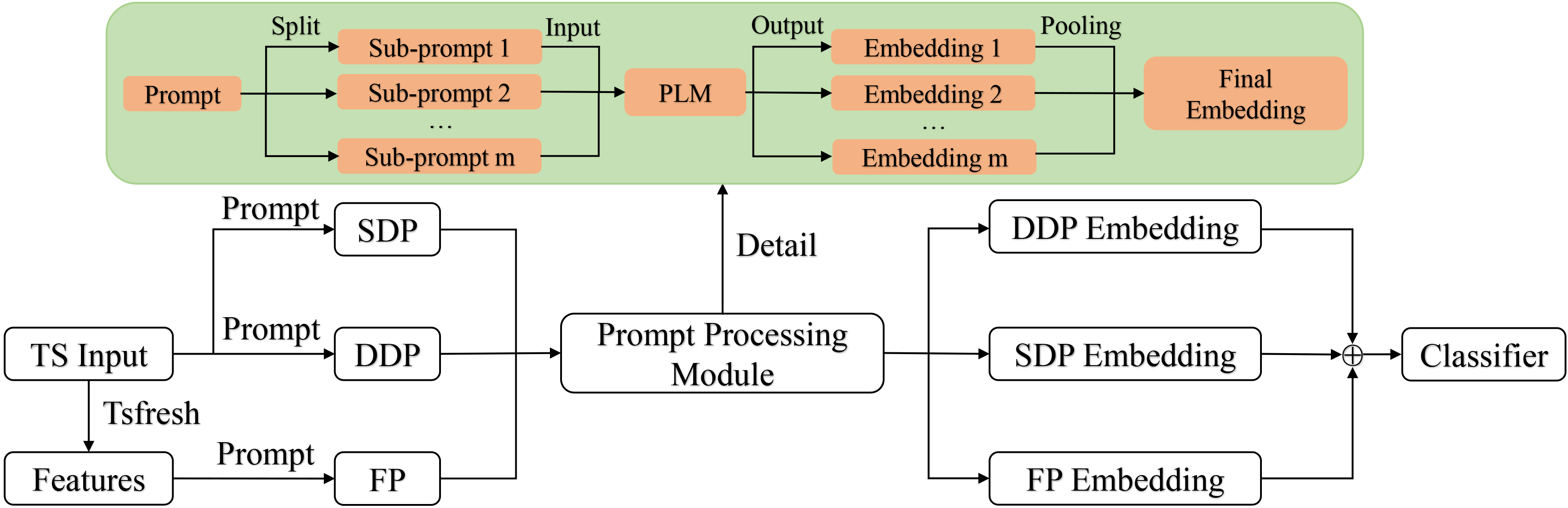}
% \end{center}
\caption{Overall pipeline of LAMPER for zero-shot time series classification.}
\label{fig1}
\end{figure}

The overall pipeline of LAMPER is illustrated in Figure~\ref{fig1}. The process begins with the design of three prompts: the Simple Description Prompt (SDP), Detailed Description Prompt (DDP), and Feature Prompt (FP), whose formats are detailed in Appendix~\ref{sa}. These prompts are strategically crafted to harness the strengths of both prompts and PLM to effectively represent features of TS data. To accommodate the maximum token input limit for the PLM, we overcome this challenge by slicing the TS into multiple sub-sequences and constructing corresponding sub-prompts. Following encoding with the PLM, we obtain multiple embeddings. The final embedding, serving as the feature representation of the TS, is acquired through a pooling method. Here, we deploy two PLMs with different length token constraints, namely the longformer \citep{Beltagy2020Longformer} with 4096 maximum token length and the BERT \citep{devlin2019bert} with 512 maximum token length. The features for the FP are obtained from the feature extraction algorithm of the Tsfresh \citep{christ2018time} module.

% The overall pipeline of LAMPER is elucidated in Figure~\ref{fig1}. We first design two distinct prompts tailored for TS analysis: the simple description prompt (SDP) and the detailed description prompt (DDP). To adhere to BERT's maximum 512-token input constraint, we partition the TS into multiple sub-series, utilizing them to populate the prompts before feeding them into BERT. The ultimate embedding output is derived through max-pooling to process the diverse outputs generated by BERT. Concurrently, Tsfresh \citep{christ2018time} is employed to extract features from the TS, which are then encapsulated as feature prompts (FP). Similar to the approach adopted for TS prompts, we slice the FP to conform to BERT's maximum input capacity. Appendix~\ref{sa} presents the prompts' format. 

\section{Experiments}
\label{exp}

\textbf{Dataset and Experimental setting.} Our study leverages a corpus of 128 univariate datasets sourced from the UCR archive \citep{8894743} to achieve the training and evaluation of LAMPER. LAMPER employs the 'bert-base-uncased' and 'longformer-base-4096' PLMs\footnote{https://huggingface.co/bert-base-uncased, https://huggingface.co/allenai/longformer-base-4096} to extract features from prompts. Additionally, we configure the Tsfresh to extract 11 features from TS, including sum, median, mean, length, standard deviation, variance, root mean square, maximum, absolute maximum, and minimum value. The final phase involves the training of a SVM classifier with the RBF kernel.

\textbf{Results}. Table~\ref{table1} delineates the outcomes obtained by employing SVM with the raw TS input, various prompts, and integration of multiple prompts (details available in appendix~\ref{sd}). Confronted with the observed inadequacy in the zero-shot classification endeavor, we meticulously scrutinize the performance of LAMPER across various datasets. Our deduction suggests that PLMs possess a constraint in grasping the nuances of TS data, despite a marginal enhancement in performance attributable to detailed prompts. Simultaneously, we posit that the imposition of a maximum token input constraint by PLMs results in the inadvertent loss of crucial contextual information embedded within the TS data, with the Longformer model therefore outperforming BERT. The Critical Sifference (CD) diagram \citep{demvsar2006statistical} for Nemenyi tests on 128 datasets is presented in Appendix~\ref{sb} and the ablation experiments results pertaining to the prompts fusion are delineated in Appendix~\ref{sc}.

\begin{table}[ht]
    \centering
    \begin{tabular}{l|l|l|l|l|l|l}
    \hline
        Method & Metrics & SDP & DDP & FP & Fusion & TS (Benchmark) \\ \hline
        BERT & Average Accuracy & 45.02\% & 48.60\% & 41.54\% & 49.39\% & 79.99\%  \\ 
        ~ & Average Rank & 8.79 & 8.77 & 10.15 & 7.65 & 2.06 \\ \hline
        Longformer & Average Accuracy & 47.50\% & 49.75\% & 48.68\% & 51.47\% & 79.99\% \\ 
        ~ & Average Rank & 8.72 & 8.62 & 8.26 & 7.29 & 2.06 \\ \hline
    \end{tabular}
    \caption{Zero-shot classification results of various prompts based on SVM.}
    \label{table1}
\end{table}

\section{Discussion and Conclusion}
\label{con}
This paper delves into an investigation into the impact of various prompts and their integration on enhancing the performance of PLMs in zero-shot TS classification tasks. Despite their adeptness in diverse zero-shot tasks within NLP, our endeavors to implement zero-shot TS classification through collaborative prompt engineering and PLMs did not yield the anticipated outcomes. In general, our experimental findings yield insightful observations: (i) The constraints imposed by the maximum input length of PLMs necessitate the segmentation of TS data, resulting in a loss of contextual information when fed into PLMs. This compromise adversely affects the feature representation of TS data, with a discernible performance decline in PLMs as the length of the TS increases in most cases. Notably, the introduction of a well-designed TS encoder proves instrumental in ameliorating PLMs' performance in zero-shot TS tasks \citep{sun2023test, liu2023pttuning}. (ii) Our results underscore the influence of diverse prompts on zero-shot classification outcomes, emphasizing that the integration of multiple prompts does not uniformly confer improvements to the model. Subsequent investigations should focus on the construction of varied prompt types, such as sentiment analysis and mask filling derived from PLMs, alongside the development of a multi-prompts fusion model. These avenues hold promise for augmenting the adaptive capabilities of PLMs in handling TS data.

\subsubsection*{Acknowledgements}
This research is supported by the National Natural Science Foundation of China 31970752,32350410397;Science,Technology,Innovation Commission of Shenzhen Municipality,JCYJ20220530143014032,JCYJ20230807113017035,WDZC20200820173710001,Shenzhen Medical Research Funds,D2301002;Department of Chemical Engineering-iBHE special cooperation joint fund project, DCE-iBHE-2022-3; Tsinghua Shenzhen Interna-tional Graduate School Cross-disciplinary Research and Innovation Fund Research Plan, JC2022009; and Bureau of Planning, Land and Resources of Shenzhen Municipality (2022) 207.

\subsubsection*{URM Statement}
The authors acknowledge that all key authors of this work meet the URM criteria of ICLR 2024 Tiny Papers Track.

\subsubsection*{Reproducibility Statement}
The source code for this work will be available at https://github.com/dodoxxb/lamper.

\bibliography{iclr2023_conference_tinypaper}
\bibliographystyle{iclr2023_conference_tinypaper}

\clearpage
\appendix
\section{Prompts' format}
\label{sa}
\begin{table}[!ht]
    \centering
    \begin{tabularx}{\textwidth}{|c|X|}
    \hline
        Prompt & Format \\ \hline
        SDP & Raw values of time series  \\ \hline
        DDP & \emph{The length of time series is [\textbf{length of time series}]. The original time series is splited into [\textbf{number of sub-series}] sub-series, whose length is [\textbf{length of sub-series}]. The specific value of the [\textbf{index of sub-series}] sub-series are [\textbf{sub-series data}] in order.}  \\ \hline
        FP & \emph{\textbf{[num of features]} features of the time series are extracted via tsfresh, the feature of \textbf{[feature name]} is \textbf{[feature value]}, ... , the feature of \textbf{[feature name]} is \textbf{[feature value]}.}  \\ \hline
    \end{tabularx}
    \caption{The format of various prompts used in this study.}
\end{table}

\section{CD diagram of LAMPER with SVM}
\label{sb}
\begin{figure}[ht]
% \begin{center}
%\framebox[4.0in]{$\;$}
\includegraphics[width=1\linewidth]{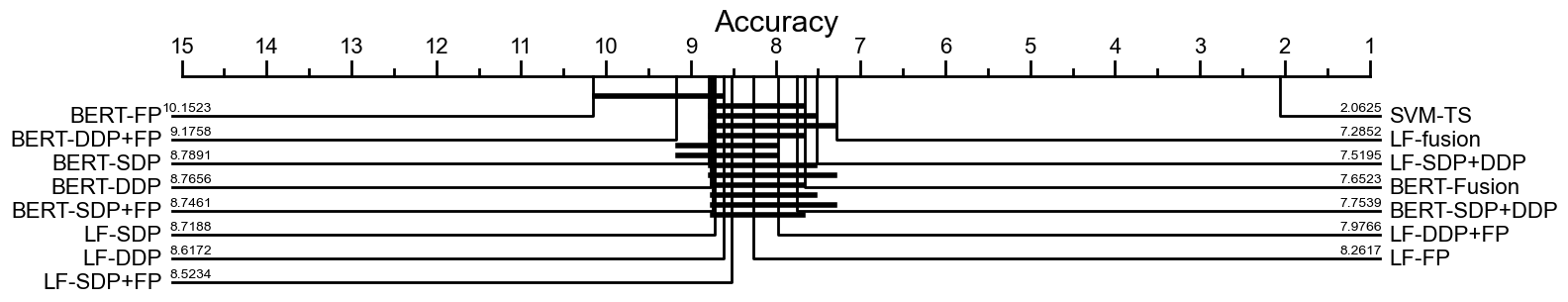}
% \end{center}
\caption{CD diagram of LAMPER with SVM on zero-shot time series classification tasks with a confidence level of 95\%, where classifiers that are not connected by a bold line are significantly different in average ranks.}
\label{afig2}
\end{figure}

\clearpage
\section{Ablation experiment}
\label{sc}
\begin{table}[ht]
    \centering
    \begin{tabular}{c|c|c|c|c|c}
    \hline
        Method & Metrics & SDP+DDP & SDP+FP & DDP+FP & Fusion \\ \hline
        BERT & Average Accuracy & 49.73\% & 45.13\% & 45.49\% & 49.39\% \\
        ~ & Average Rank & 7.75 & 8.75 & 9.18 & 7.65\\ \hline
        Longformer & Average Accuracy & 51.12\% & 47.88\% & 50.96\% & 51.47\% \\ 
        ~ & Average Rank & 7.52 & 8.52 & 7.98 & 7.29 \\ \hline
    \end{tabular}
    \caption{Ablation experiment results of prompts fusion.}
\end{table}

\section{128 univariate datasets of UCR archive and experimental results}
\label{sd}
% You may include other additional sections here. However, please be mindful that the spirit of the Tiny Papers track is for papers to be short. Avoid overly-long appendices.
\begin{longtable}{p{0.9cm}|p{0.9cm}p{0.9cm}p{0.9cm}p{1cm}|p{0.9cm}p{0.9cm}p{0.9cm}p{1cm}|p{0.9cm}}
\caption{Description of 128 univariate datasets of UCR archive and detailed results of PLMs}\\
    % \centering
    % \begin{tabular}
    \hline
        ~ & BERT & ~ & ~ & ~ & Longformer & ~ & ~ & ~ & ~ \\ \hline
        Length & SVM-SDP & SVM-DDP & SVM-FP & SVM-Fusion & SVM-SDP & SVM-DDP & SVM-FP & SVM-Fusion & SVM-TS \\ \hline
        1460 & 34.0\% & 60.0\% & 29.0\% & 47.0\% & 70.0\% & 72.0\% & 65.0\% & 79.0\% & 58.0\% \\ \hline
        176 & 3.8\% & 3.8\% & 3.8\% & 3.8\% & 3.8\% & 3.8\% & 3.8\% & 3.8\% & 9.2\% \\ \hline
        Vary & 27.3\% & 68.7\% & 21.0\% & 36.3\% & 32.3\% & 72.3\% & 34.0\% & 51.3\% & 66.3\% \\ \hline
        Vary & 30.7\% & 69.3\% & 21.3\% & 39.0\% & 26.0\% & 70.3\% & 41.7\% & 49.3\% & 67.3\% \\ \hline
        Vary & 26.3\% & 62.3\% & 17.3\% & 23.7\% & 36.0\% & 67.7\% & 40.3\% & 57.0\% & 54.0\% \\ \hline
        251 & 63.9\% & 36.1\% & 52.8\% & 63.9\% & 66.7\% & 36.1\% & 77.8\% & 72.2\% & 75.0\% \\ \hline
        470 & 53.3\% & 73.3\% & 53.3\% & 60.0\% & 46.7\% & 70.0\% & 50.0\% & 83.3\% & 50.0\% \\ \hline
        512 & 60.0\% & 70.0\% & 65.0\% & 75.0\% & 65.0\% & 95.0\% & 75.0\% & 100.0\% & 100.0\% \\ \hline
        512 & 70.0\% & 55.0\% & 65.0\% & 80.0\% & 80.0\% & 70.0\% & 70.0\% & 90.0\% & 90.0\% \\ \hline
        128 & 76.7\% & 80.0\% & 60.0\% & 73.3\% & 86.7\% & 73.3\% & 73.3\% & 73.3\% & 80.0\% \\ \hline
        577 & 28.3\% & 28.3\% & 28.3\% & 28.3\% & 28.3\% & 28.3\% & 28.3\% & 28.3\% & 55.0\% \\ \hline
        128 & 40.0\% & 40.0\% & 40.0\% & 40.0\% & 40.0\% & 40.0\% & 40.0\% & 40.0\% & 96.7\% \\ \hline
        24 & 70.0\% & 70.0\% & 55.0\% & 80.0\% & 80.0\% & 75.0\% & 70.0\% & 80.0\% & 80.0\% \\ \hline
        166 & 56.1\% & 56.1\% & 56.1\% & 56.1\% & 56.1\% & 56.1\% & 56.1\% & 56.1\% & 58.0\% \\ \hline
        1639 & 32.5\% & 32.5\% & 32.5\% & 32.5\% & 32.5\% & 32.5\% & 32.5\% & 32.5\% & 100.0\% \\ \hline
        286 & 57.1\% & 82.1\% & 71.4\% & 75.0\% & 71.4\% & 75.0\% & 89.3\% & 82.1\% & 96.4\% \\ \hline
        720 & 62.0\% & 55.6\% & 56.0\% & 70.4\% & 59.2\% & 55.6\% & 70.0\% & 60.0\% & 90.4\% \\ \hline
        300 & 10.0\% & 10.0\% & 10.0\% & 10.0\% & 10.0\% & 10.0\% & 10.0\% & 10.0\% & 87.9\% \\ \hline
        300 & 10.5\% & 9.7\% & 13.6\% & 12.6\% & 10.3\% & 10.3\% & 12.8\% & 10.3\% & 80.8\% \\ \hline
        300 & 10.5\% & 10.5\% & 10.5\% & 10.5\% & 10.5\% & 10.5\% & 10.5\% & 10.5\% & 85.6\% \\ \hline
        46 & 22.9\% & 70.9\% & 6.5\% & 28.6\% & 35.7\% & 75.9\% & 18.2\% & 54.2\% & 71.8\% \\ \hline
        345 & 37.5\% & 37.5\% & 37.5\% & 37.5\% & 37.5\% & 37.5\% & 37.5\% & 37.5\% & 37.5\% \\ \hline
        80 & 67.5\% & 64.8\% & 64.3\% & 65.8\% & 64.3\% & 64.3\% & 64.3\% & 64.3\% & 80.8\% \\ \hline
        80 & 63.0\% & 63.0\% & 63.0\% & 63.0\% & 63.0\% & 63.0\% & 63.0\% & 63.0\% & 71.5\% \\ \hline
        80 & 70.8\% & 53.3\% & 53.3\% & 53.3\% & 53.3\% & 53.3\% & 53.3\% & 53.3\% & 76.8\% \\ \hline
        288 & 16.7\% & 39.7\% & 21.8\% & 30.8\% & 28.2\% & 39.7\% & 39.7\% & 32.1\% & 69.2\% \\ \hline
        288 & 60.0\% & 75.0\% & 70.0\% & 75.0\% & 55.0\% & 65.0\% & 85.0\% & 80.0\% & 100.0\% \\ \hline
        288 & 55.0\% & 95.0\% & 55.0\% & 95.0\% & 95.0\% & 100.0\% & 90.0\% & 100.0\% & 100.0\% \\ \hline
        512 & 82.0\% & 82.0\% & 82.0\% & 82.0\% & 82.0\% & 82.0\% & 82.0\% & 82.0\% & 92.9\% \\ \hline
        96 & 69.0\% & 69.0\% & 69.0\% & 69.0\% & 69.0\% & 69.0\% & 69.0\% & 69.0\% & 87.0\% \\ \hline
        140 & 75.4\% & 58.4\% & 58.4\% & 72.2\% & 58.4\% & 58.4\% & 58.4\% & 58.4\% & 95.0\% \\ \hline
        136 & 60.9\% & 60.9\% & 60.9\% & 60.9\% & 60.9\% & 60.9\% & 60.9\% & 60.9\% & 82.6\% \\ \hline
        96 & 53.5\% & 32.9\% & 27.0\% & 51.6\% & 45.5\% & 30.5\% & 27.0\% & 44.5\% & 84.7\% \\ \hline
        1250 & 26.8\% & 10.5\% & 13.0\% & 20.4\% & 13.8\% & 9.9\% & 14.9\% & 13.5\% & 71.0\% \\ \hline
        1250 & 17.4\% & 8.6\% & 13.0\% & 14.6\% & 12.7\% & 8.6\% & 12.4\% & 12.7\% & 51.1\% \\ \hline
        1751 & 31.2\% & 27.2\% & 30.6\% & 33.7\% & 36.5\% & 29.2\% & 30.8\% & 34.5\% & 33.1\% \\ \hline
        131 & 26.4\% & 71.3\% & 16.6\% & 31.1\% & 26.8\% & 75.2\% & 30.4\% & 42.3\% & 98.8\% \\ \hline
        350 & 50.0\% & 37.5\% & 37.5\% & 54.2\% & 54.2\% & 37.5\% & 66.7\% & 50.0\% & 95.8\% \\ \hline
        131 & 16.5\% & 16.5\% & 16.5\% & 16.5\% & 16.5\% & 16.5\% & 16.5\% & 16.5\% & 96.0\% \\ \hline
        270 & 11.6\% & 11.6\% & 11.6\% & 11.6\% & 11.6\% & 11.6\% & 11.6\% & 11.6\% & 77.1\% \\ \hline
        463 & 25.7\% & 17.7\% & 16.6\% & 19.4\% & 25.1\% & 18.9\% & 22.9\% & 25.7\% & 53.1\% \\ \hline
        500 & 51.3\% & 51.3\% & 51.3\% & 51.3\% & 51.3\% & 51.3\% & 51.3\% & 51.3\% & 94.0\% \\ \hline
        500 & 51.6\% & 51.1\% & 51.2\% & 51.2\% & 51.2\% & 51.2\% & 51.2\% & 51.2\% & 95.9\% \\ \hline
        301 & 70.7\% & 59.3\% & 71.3\% & 70.7\% & 67.3\% & 55.3\% & 79.3\% & 84.7\% & 78.0\% \\ \hline
        301 & 85.7\% & 67.9\% & 78.6\% & 89.3\% & 71.4\% & 71.4\% & 100.0\% & 96.4\% & 85.7\% \\ \hline
        201 & 100.0\% & 100.0\% & 100.0\% & 100.0\% & 100.0\% & 100.0\% & 100.0\% & 100.0\% & 100.0\% \\ \hline
        Vary & 35.6\% & 6.7\% & 23.1\% & 47.1\% & 34.1\% & 8.2\% & 30.3\% & 36.5\% & 68.3\% \\ \hline
        Vary & 41.3\% & 7.2\% & 14.4\% & 31.3\% & 45.2\% & 7.2\% & 24.0\% & 51.4\% & 54.8\% \\ \hline
        Vary & 30.3\% & 9.6\% & 15.9\% & 35.1\% & 14.4\% & 9.6\% & 29.8\% & 14.9\% & 40.4\% \\ \hline
        Vary & 30.3\% & 18.9\% & 18.9\% & 18.9\% & 18.9\% & 18.9\% & 18.9\% & 18.9\% & 100.0\% \\ \hline
        Vary & 31.5\% & 17.8\% & 17.8\% & 17.8\% & 17.8\% & 17.8\% & 17.8\% & 17.8\% & 97.9\% \\ \hline
        150 & 52.0\% & 52.0\% & 52.0\% & 52.0\% & 52.0\% & 52.0\% & 52.0\% & 52.0\% & 88.0\% \\ \hline
        150 & 50.4\% & 72.6\% & 50.4\% & 51.1\% & 50.4\% & 50.4\% & 50.4\% & 50.4\% & 90.4\% \\ \hline
        150 & 52.6\% & 52.6\% & 52.6\% & 52.6\% & 52.6\% & 52.6\% & 52.6\% & 52.6\% & 97.0\% \\ \hline
        150 & 88.2\% & 52.2\% & 52.2\% & 66.2\% & 52.2\% & 52.2\% & 52.2\% & 52.2\% & 100.0\% \\ \hline
        431 & 52.3\% & 52.3\% & 52.3\% & 52.3\% & 52.3\% & 52.3\% & 52.3\% & 52.3\% & 90.8\% \\ \hline
        2709 & 63.8\% & 63.8\% & 63.8\% & 63.8\% & 63.8\% & 63.8\% & 63.8\% & 63.8\% & 86.4\% \\ \hline
        1092 & 23.2\% & 23.2\% & 23.2\% & 23.2\% & 23.2\% & 23.2\% & 23.2\% & 23.2\% & 57.4\% \\ \hline
        512 & 60.9\% & 60.9\% & 60.9\% & 60.9\% & 60.9\% & 60.9\% & 60.9\% & 60.9\% & 60.9\% \\ \hline
        2000 & 67.5\% & 57.5\% & 55.0\% & 92.5\% & 82.5\% & 67.5\% & 55.0\% & 95.0\% & 100.0\% \\ \hline
        1882 & 23.0\% & 19.0\% & 25.0\% & 24.0\% & 27.0\% & 22.0\% & 29.0\% & 27.0\% & 43.0\% \\ \hline
        601 & 100.0\% & 48.4\% & 83.9\% & 83.9\% & 48.4\% & 48.4\% & 48.4\% & 48.4\% & 100.0\% \\ \hline
        601 & 52.9\% & 47.1\% & 47.1\% & 47.1\% & 47.1\% & 47.1\% & 47.1\% & 47.1\% & 100.0\% \\ \hline
        256 & 13.2\% & 15.0\% & 16.8\% & 35.0\% & 31.4\% & 15.0\% & 48.2\% & 35.9\% & 74.5\% \\ \hline
        24 & 50.7\% & 50.7\% & 50.7\% & 50.7\% & 50.7\% & 50.7\% & 50.7\% & 50.7\% & 98.5\% \\ \hline
        720 & 43.2\% & 59.7\% & 42.9\% & 43.2\% & 57.1\% & 62.7\% & 56.5\% & 64.5\% & 90.7\% \\ \hline
        637 & 66.7\% & 66.7\% & 66.7\% & 66.7\% & 66.7\% & 66.7\% & 66.7\% & 66.7\% & 85.0\% \\ \hline
        319 & 27.1\% & 27.1\% & 27.1\% & 27.1\% & 27.1\% & 27.1\% & 27.1\% & 27.1\% & 88.6\% \\ \hline
        1024 & 20.0\% & 20.0\% & 20.0\% & 20.0\% & 20.0\% & 20.0\% & 20.0\% & 20.0\% & 72.7\% \\ \hline
        448 & 45.0\% & 43.3\% & 40.0\% & 43.3\% & 50.0\% & 65.0\% & 71.7\% & 73.3\% & 96.7\% \\ \hline
        99 & 53.3\% & 53.3\% & 53.3\% & 53.3\% & 53.3\% & 53.3\% & 53.3\% & 53.3\% & 64.6\% \\ \hline
        24 & 35.3\% & 65.8\% & 18.9\% & 50.0\% & 41.5\% & 63.7\% & 26.0\% & 52.7\% & 84.6\% \\ \hline
        80 & 67.3\% & 62.8\% & 59.3\% & 60.0\% & 59.3\% & 59.3\% & 59.3\% & 59.3\% & 77.3\% \\ \hline
        80 & 64.7\% & 64.7\% & 64.7\% & 64.7\% & 64.7\% & 64.7\% & 64.7\% & 64.7\% & 64.7\% \\ \hline
        80 & 60.4\% & 40.1\% & 40.1\% & 40.1\% & 40.1\% & 40.1\% & 40.1\% & 40.1\% & 61.2\% \\ \hline
        1024 & 28.4\% & 73.4\% & 38.4\% & 52.6\% & 47.2\% & 72.6\% & 37.2\% & 52.4\% & 91.4\% \\ \hline
        1024 & 37.0\% & 58.0\% & 37.0\% & 68.0\% & 55.0\% & 74.0\% & 58.0\% & 67.0\% & 93.0\% \\ \hline
        84 & 60.0\% & 60.0\% & 75.0\% & 90.0\% & 60.0\% & 60.0\% & 70.0\% & 65.0\% & 95.0\% \\ \hline
        750 & 3.4\% & 3.1\% & 5.3\% & 3.1\% & 3.1\% & 3.1\% & 3.1\% & 3.1\% & 66.0\% \\ \hline
        750 & 3.1\% & 3.1\% & 5.1\% & 3.1\% & 3.1\% & 3.1\% & 3.1\% & 3.1\% & 70.4\% \\ \hline
        570 & 43.3\% & 43.3\% & 43.3\% & 43.3\% & 43.3\% & 43.3\% & 43.3\% & 43.3\% & 43.3\% \\ \hline
        427 & 26.5\% & 26.5\% & 26.5\% & 26.5\% & 26.5\% & 26.5\% & 26.5\% & 26.5\% & 85.5\% \\ \hline
        80 & 65.1\% & 65.1\% & 65.1\% & 65.1\% & 65.1\% & 65.1\% & 65.1\% & 65.1\% & 67.1\% \\ \hline
        1024 & 11.2\% & 11.2\% & 11.2\% & 11.2\% & 11.2\% & 11.2\% & 11.2\% & 11.2\% & 78.0\% \\ \hline
        Vary & 56.0\% & 70.0\% & 30.0\% & 54.0\% & 76.0\% & 86.0\% & 68.0\% & 92.0\% & 94.0\% \\ \hline
        2000 & 39.4\% & 71.2\% & 56.7\% & 79.8\% & 91.3\% & 85.6\% & 98.1\% & 97.1\% & 43.3\% \\ \hline
        2000 & 41.3\% & 76.9\% & 47.1\% & 58.7\% & 79.8\% & 86.5\% & 95.2\% & 85.6\% & 93.3\% \\ \hline
        2000 & 37.5\% & 55.8\% & 38.5\% & 66.3\% & 80.8\% & 79.8\% & 92.3\% & 86.5\% & 90.4\% \\ \hline
        Vary & 16.9\% & 31.3\% & 28.7\% & 30.9\% & 16.4\% & 16.4\% & 16.4\% & 16.4\% & 30.5\% \\ \hline
        144 & 19.0\% & 19.0\% & 19.0\% & 19.0\% & 19.0\% & 19.0\% & 19.0\% & 19.0\% & 99.0\% \\ \hline
        144 & 79.4\% & 60.0\% & 67.8\% & 82.8\% & 70.0\% & 56.1\% & 78.9\% & 75.0\% & 100.0\% \\ \hline
        80 & 73.0\% & 74.3\% & 47.3\% & 73.3\% & 47.3\% & 47.3\% & 47.3\% & 47.3\% & 74.8\% \\ \hline
        80 & 67.7\% & 67.7\% & 67.7\% & 67.7\% & 67.7\% & 67.7\% & 67.7\% & 67.7\% & 67.8\% \\ \hline
        80 & 68.8\% & 45.0\% & 45.0\% & 71.3\% & 45.0\% & 45.0\% & 45.0\% & 45.0\% & 72.3\% \\ \hline
        720 & 44.0\% & 66.4\% & 36.8\% & 53.3\% & 45.1\% & 70.9\% & 42.1\% & 58.7\% & 99.2\% \\ \hline
        2844 & 55.0\% & 85.0\% & 50.0\% & 85.0\% & 80.0\% & 80.0\% & 95.0\% & 90.0\% & 85.0\% \\ \hline
        720 & 43.5\% & 68.3\% & 37.6\% & 48.3\% & 45.9\% & 61.3\% & 49.3\% & 58.1\% & 82.7\% \\ \hline
        1500 & 53.7\% & 66.0\% & 53.0\% & 62.3\% & 72.7\% & 52.7\% & 53.3\% & 55.0\% & 94.0\% \\ \hline
        1500 & 24.4\% & 51.6\% & 21.6\% & 47.3\% & 43.3\% & 75.3\% & 42.2\% & 60.2\% & 66.2\% \\ \hline
        1500 & 27.6\% & 53.1\% & 28.9\% & 51.3\% & 33.6\% & 66.4\% & 40.2\% & 47.8\% & 83.6\% \\ \hline
        Vary & 52.0\% & 62.0\% & 28.0\% & 56.0\% & 76.0\% & 84.0\% & 72.0\% & 92.0\% & 86.0\% \\ \hline
        500 & 55.0\% & 95.0\% & 65.0\% & 90.0\% & 60.0\% & 75.0\% & 85.0\% & 85.0\% & 100.0\% \\ \hline
        512 & 6.3\% & 66.8\% & 12.5\% & 37.3\% & 32.5\% & 80.3\% & 33.0\% & 43.8\% & 83.3\% \\ \hline
        720 & 46.4\% & 74.9\% & 43.2\% & 59.7\% & 53.1\% & 61.6\% & 55.2\% & 71.7\% & 98.4\% \\ \hline
        15 & 44.7\% & 54.0\% & 40.7\% & 54.0\% & 62.0\% & 73.3\% & 75.3\% & 72.0\% & 98.7\% \\ \hline
        70 & 70.0\% & 70.0\% & 70.0\% & 70.0\% & 70.0\% & 70.0\% & 70.0\% & 70.0\% & 90.0\% \\ \hline
        65 & 59.3\% & 59.3\% & 59.3\% & 59.3\% & 59.3\% & 59.3\% & 59.3\% & 59.3\% & 92.6\% \\ \hline
        1024 & 57.3\% & 57.3\% & 70.0\% & 78.7\% & 57.3\% & 57.3\% & 57.3\% & 57.3\% & 84.8\% \\ \hline
        235 & 64.3\% & 64.3\% & 64.3\% & 64.3\% & 64.3\% & 64.3\% & 64.3\% & 64.3\% & 64.4\% \\ \hline
        128 & 8.4\% & 8.4\% & 8.4\% & 8.4\% & 8.4\% & 8.4\% & 8.4\% & 8.4\% & 80.4\% \\ \hline
        398 & 32.0\% & 32.0\% & 32.0\% & 32.0\% & 32.0\% & 32.0\% & 32.0\% & 32.0\% & 80.0\% \\ \hline
        60 & 34.0\% & 69.7\% & 21.7\% & 73.0\% & 58.0\% & 78.0\% & 30.0\% & 71.7\% & 100.0\% \\ \hline
        277 & 60.0\% & 50.0\% & 65.0\% & 57.5\% & 60.0\% & 52.5\% & 87.5\% & 62.5\% & 97.5\% \\ \hline
        343 & 55.6\% & 55.6\% & 61.1\% & 72.2\% & 77.8\% & 52.8\% & 83.3\% & 72.2\% & 100.0\% \\ \hline
        275 & 31.0\% & 31.0\% & 33.0\% & 31.0\% & 35.0\% & 31.0\% & 31.0\% & 31.0\% & 69.0\% \\ \hline
        82 & 52.2\% & 52.2\% & 52.2\% & 52.2\% & 52.2\% & 52.2\% & 52.2\% & 52.2\% & 69.6\% \\ \hline
        128 & 27.1\% & 27.1\% & 29.0\% & 27.1\% & 27.1\% & 27.1\% & 27.1\% & 27.1\% & 99.6\% \\ \hline
        150 & 44.4\% & 72.2\% & 50.0\% & 66.7\% & 63.9\% & 75.0\% & 86.1\% & 69.4\% & 72.2\% \\ \hline
        945 & 20.4\% & 14.2\% & 15.3\% & 14.2\% & 14.2\% & 14.2\% & 14.2\% & 14.2\% & 98.4\% \\ \hline
        315 & 16.2\% & 14.2\% & 14.2\% & 14.2\% & 14.2\% & 14.2\% & 14.2\% & 14.2\% & 82.6\% \\ \hline
        315 & 23.3\% & 14.2\% & 18.6\% & 16.0\% & 26.9\% & 14.2\% & 14.2\% & 14.2\% & 78.7\% \\ \hline
        315 & 20.9\% & 14.2\% & 14.2\% & 14.8\% & 26.3\% & 14.2\% & 14.2\% & 14.2\% & 78.1\% \\ \hline
        152 & 90.3\% & 90.3\% & 90.3\% & 90.3\% & 90.3\% & 90.3\% & 90.3\% & 90.3\% & 99.8\% \\ \hline
        234 & 64.9\% & 52.6\% & 52.6\% & 52.6\% & 52.6\% & 52.6\% & 52.6\% & 52.6\% & 52.6\% \\ \hline
        270 & 22.5\% & 22.5\% & 22.5\% & 22.5\% & 22.5\% & 22.5\% & 22.5\% & 22.5\% & 75.7\% \\ \hline
        900 & 42.0\% & 42.0\% & 42.0\% & 42.0\% & 42.0\% & 42.0\% & 42.0\% & 42.0\% & 84.5\% \\ \hline
        900 & 58.0\% & 58.0\% & 58.0\% & 58.0\% & 58.0\% & 58.0\% & 58.0\% & 58.0\% & 90.1\% \\ \hline
        426 & 54.3\% & 54.3\% & 54.3\% & 54.3\% & 54.3\% & 54.3\% & 54.3\% & 54.3\% & 72.7\% \\ \hline
    % \end{tabular}
\end{longtable}
\end{document}